%% file: main.tex
\documentclass[10pt,twocolumn,letterpaper]{article}

\usepackage[pagenumbers]{cvpr} 
\usepackage[none]{hyphenat}
\usepackage{verbatim}
\usepackage{multirow}
\usepackage{nidanfloat}

\input{preamble}

\definecolor{cvprblue}{rgb}{0.21,0.49,0.74}
\usepackage[pagebackref,breaklinks,colorlinks,citecolor=cvprblue]{hyperref}

\usepackage{natbib}
\renewcommand\harvardurl[1]{\textbf{URL:} \url{#1}}
\usepackage{hyperref}

\def\paperID{42} 
\def\confName{CVPR}
\def\confYear{2024}

\title{Importance of realism in procedurally-generated synthetic images for deep learning: case studies in maize and canola}

\author{Nazifa Azam Khan$^1$, Mikolaj Cieslak$^2$, Ian McQuillan$^1$\\
{\tt\small nazifa.khan@usask.ca}, {\tt\small msciesla@ucalgary.ca}, {\tt\small mcquillan@cs.usask.ca} \\
$^1$Department of Computer Science, University of Saskatchewan, Saskatchewan, Saskatoon, SK, Canada\\
$^2$Department of Computer Science, University of Calgary, Calgary, Alberta, Canada}

\begin{document}
\maketitle

\begin{abstract}
Artificial neural networks are often used to identify features of crop plants. However, training their models requires many annotated images, which can be expensive and time-consuming to acquire. Procedural models of plants, such as those developed with Lindenmayer-systems (L-systems) can be created to produce visually realistic simulations, and hence images of plant simulations, where annotations are implicitly known. These synthetic images can either augment or completely replace real images in training neural networks for phenotyping tasks. In this paper, we systematically vary amounts of real and synthetic images used for training in both maize and canola to better understand situations where synthetic images generated from L-systems can help prediction on real images. This work also explores the degree to which realism in the synthetic images improves prediction. We have five different variants of a procedural canola model (these variants were created by tuning the realism while using calibration), and the deep learning results showed how drastically these results improve as the canola synthetic images are made to be more realistic. Furthermore, we see  how neural network predictions can be used to help calibrate L-systems themselves, creating a feedback loop. \end{abstract}

\section{Introduction}
\label{sec:intro}

The quality and quantity of agricultural outputs depends on many factors such as crop genotype, disease, growth patterns, nutritional deficiency, and environmental conditions. The assessment and determination of plant growth, morphology, function, composition, disease detection, or any phenotype, is collectively called plant phenotyping. Many of these factors are evident from consistent monitoring, which has been found to be crucial data towards helping with the successful cultivation of new crops. This has traditionally been done for massive numbers of plants in a manual fashion by plant breeders, however this is both time-consuming and often overly reliant on intuition \cite{plant-breeding-1}. However,  the potential of using automated high-throughput phenotyping has long been recognised as an important step forward \cite{intro_1}, and is now in the initial stages of being used to help cultivate new crop varieties \cite{article, high2019}. Furthermore, this will become even more important with increasingly extreme climate condition.

Image-based plant phenotyping is important and is gaining in popularity as a method to automatically extract useful information from plant images \cite{plant}  and to identify phenotypic traits throughout a plant’s life \cite{crisp}. One possible method to identify many features including plant architecture is to use image processing techniques  to segment the plant from the image, and then to use skeletonization techniques to identify different plant components. However, accurate skeleton extraction is challenging due to the complex geometry of plants and self-occlusion in the images \cite{ayan}. Moreover, the identification and removal of extra branches that can arise from noise in images  \cite{cai} is also difficult.  Another increasing useful approach to analyze the large volumes of data generated by phenotyping platforms is to use machine learning techniques. Artificial neural networks (ANNs) and deep learning are providing superlative results on many data and image analysis tasks \cite{leveraging}, and have successfully been used in image classification, multi-instance detection, and segmentation \cite{deep-1}. ANNs have shown better accuracy  than  image processing algorithms in certain complex image-based plant phenotyping tasks, such as leaf counting, age estimation, mutant classification \cite{jordan-ian}, plant disease detection and diagnosis from images \cite{deep-2}, classification of fruits and other organs \cite{deep-3}, and pixel-wise localization of root and shoot tips \cite{deep-4}. These are classified as regression (counting) or identification (detection or classification)  problems, and various types of ANNs with complex architectures have been implemented to solve them. 

One of the most popular and powerful networks for deep learning is the convolutional neural network (CNN), which have layers of neurons representing image pixels, and connections between layers that  perform linear filtering. They are capable of learning highly discriminative features during the training phase, and can classify plants without needing segmentation, or  feature extraction. Many works in the literature provide examples of different plants where important phenotypic traits have been extracted from plant images with them; e.g.\ automatic joint feature and classifier learning for temporal phenotype/genotype classification \cite{deep-2018}, crop lodging detection \cite{mark-sara}, and leaf counting \cite{aich}. Deep Plant \cite{deep-plant}  uses CNNs  to learn feature representations from leaves using 44  plant species.   Deep Plant Phenomics (DPP) \cite{jordan-ian} has pre-trained CNNs built using TensorFlow for  common phenotyping tasks involving object detection, object counting, and  semantic segmentation. 

Supervised machine learning methods require a training set where correct labels for the phenotype of interest are known, and the model is trained on the labelled dataset. A major challenge for deep learning applications for plant phenotyping is the availability of a large quantity of annotated data for training. Image datasets of the desired species, environment, scale, and size, labelled with the phenotypic properties of interest, are likely not available, and might be difficult to obtain due to the large cost of collecting and annotating data \cite{jordan}. Hence, a motivation arises to use an alternative source of training data.

Computational systems for modelling and simulating plants have been an important area of research. Such simulations can incorporate different plant geometries \cite{pru, kol, quantifying}, environmental factors \cite{using-l, rice}, and mechanistic controls \cite{kol, control}. One of the most common systems used for this purpose are Lindenmayer systems (L-systems), introduced by Lindenmayer in 1968 \cite{lind} as a formalism for simulating the development of multicellular organisms in terms of division, growth, and death of individual cells. They can be used as a mathematical theory of plant development  \cite{pru} and for simulating plants  using a simulator such as \cite{vlab}. 

The idea of using synthetic images from these simulations to train ANNs for the purpose of identifying phenotypic properties was initially explored by Ubbens et al.\  \cite{jordan}. This idea overcomes many of the obstacles of manual annotations. If the phenotype of interest is built into the L-system model, then the correct annotations are automatically known. For the purposes of identifying plant organs and architecture, these can be precisely incorporated into an L-system model  \cite{pru,  l-2}. In comparison to the cost of real data collection, and the precise labelling of them, it is easier to obtain synthetic data from these procedural models. Once a model has been created for a species, it is possible to generate arbitrarily large synthetic images datasets at no extra expense. They showed \cite{jordan} that synthetic images from an L-system can be used to augment datasets of real plant images or can even be used alone as a large source of training data. Using both together led to better prediction of leaf count in \textit{Arabidopsis thaliana}. A similar approach was also used with maize to predict leaf count \cite{ian-jordan-maize} but using a maize procedural model within Plant Factory Exporter \cite{plant_factory} instead of L-systems to create synthetic images. Here, results were mixed and there were cases (depending on the number of synthetic images used) where adding synthetic images to a real dataset for training improved prediction versus only training on real images, but often adding synthetic images did not help.

Cieslak et al.\ \cite{mik-nazifa-lsystem} demonstrated an approach for creating L-system models using real images for guidance, for maize and canola as a case study. 
Primarily, the modelling process has two phases: the first is the qualitative or topological specification of the developing plant architecture, and the second is the  calibration, so they can accurately capture the growth and visual characteristics in order to generate realistic synthetic images. Indeed, one might expect that the more realistic the synthetic images look, the more replaceable they are for real images for training. But the degree to which this is true is not yet understood. Their approach also demonstrates that it is particularly easy to take an existing L-system model and adapt it to a new environment, which is another big potential advantage of using images from L-systems over manually labelling in each new environment. 

In this paper, we continue studying situations where it may or may not be helpful to use  synthetic images for training ANNs. We use  two complex plants, maize and canola, and start with the same L-systems created in \cite{mik-nazifa-lsystem}. 

Maize is an important cereal crop, with thin leaves, leaf occlusion, and large leaf curvatures, that make them difficult to analyze. We systematically vary different amounts of real images together with synthetic images from L-system simulations, and measure success at leaf counting on real images using both mean absolute loss and Pearson's correlation. We find that using synthetic images in addition to real images always improves success, with a large benefit when the number of real images is small, and the benefit decreasing as the number of real images increases.

The next experiment involves canola, which is even more complex in our dataset with larger numbers of occluded leaves. We try to predict the inflorescence branch numbers. Here, ANNs trained with only synthetic images and no real images work reasonably well at predicting inflorescence branch numbers in real images. However, in most situations real annotated images work better for training than a combination of real and synthetic images. While visually these synthetic images look similar to real plant images, refining the L-system models for the purposes of creating even more realistic synthetic images has a considerable benefit. The accuracy of the predictions was quite sensitive to the calibration of the L-systems, and improving the realism of the synthetic images had a substantial effect on the predictive accuracy. The ANN prediction results were used to help improve the L-system calibration process, which were in-turn used to create even more accurate L-systems and synthetic images. This process can improve both the phenotype predictions, but also the L-systems themselves.

\section{Dataset}
\label{sec:data}

\paragraph{Maize dataset:} Real images of maize were obtained from an open dataset created at the University of Nebraska-Lincoln \cite{dataset}. The dataset (called UNL-CPPD-I) contains 700 total images of maize plants taken using the visible light camera at the UNL Lemnatec Scanalyzer 3D High-Throughput Phenotyping Facility \cite{unl}. Images were taken of 13 maize plants with different genotypes for 27 consecutive days of development during the vegetative stage starting 2 days after seed planting. Images were taken from two different side views separated by 90 degrees. The dataset also contains ground-truth annotations where leaves that are visible in the images are marked (leaves that were completely occluded in the images were not annotated). This is immediately evident because the number of annotated leaves from the two views can differ substantially. While this is advantageous from the perspective of identifying leaves on an individual image, it does hinder the evaluation of leaf identification procedures that try to identify leaves even if they are occluded, which is also interesting to know. Additional details regarding the imaging setup, dataset organization, and genotypes are provided in \cite{unl}. 

Maize follows an alternate phyllotaxy with each leaf developing on the opposite side of the previous leaf, and therefore the leaves form a planar-like surface. A pre-processing step was used to determine the best view between the two available views. The same procedure was performed as described in \cite{nazifa}. Briefly, background subtraction was used to extract the foreground to remove the fixed background of the phenotyping system, Otsu thresholding was used on the grayscale image of the foreground image to obtain the segmented image, and another thresholding was performed by calculating the excess green index of the image. At this point, for each plant and each day, the convex hulls of the binarized plant images of both views were calculated, and the one with the largest convex hull was selected as the view for that plant/day. Only the resulting images chosen via this procedure were used for all subsequent analyses. Lastly, only images with at least one leaf were kept, and this dataset is known as the real maize dataset, denoted by $M^R$. In total, this dataset contains 328 images.

\begin{figure}[!htbp]
  \centering
  \subfloat[]{\includegraphics[width=0.4\linewidth]{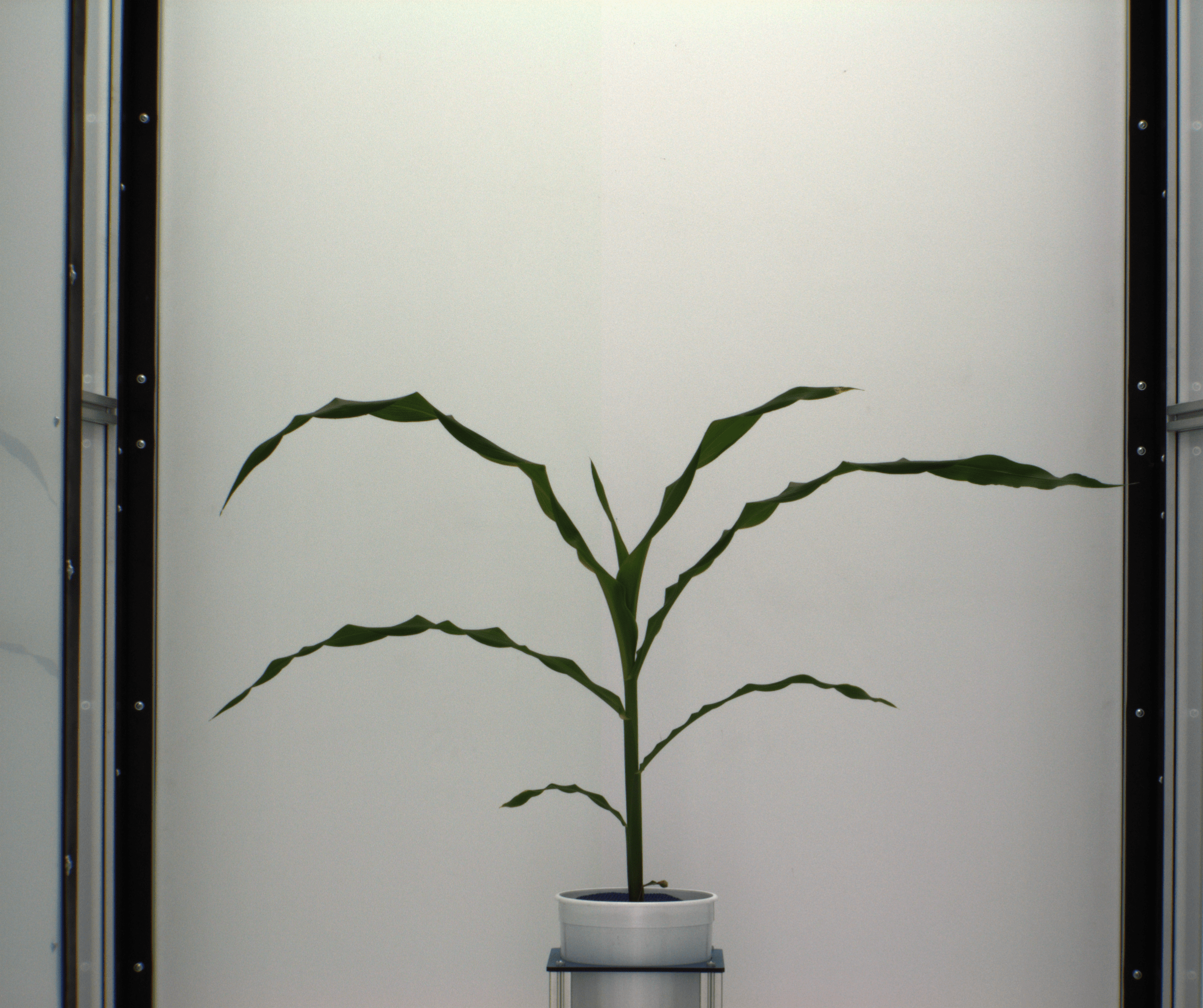}\label{fig_met_8}}
  \hspace{1em} 
  \subfloat[]{\includegraphics[width=0.4\linewidth]{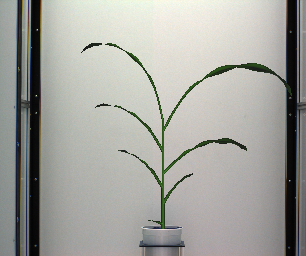}\label{fig_met_9}}
   \caption{(a) A real maize image on day 25 from 0 degree view. (b) A synthetic maize plant on day 25 generated from the maize L-system.}
  \label{fig_met_11}
\end{figure}

In total, the development of 50 plants was simulated from the maize L-system where they were calibrated to use the same timeline as the real plants. However, only a single view was captured for each day and plant, taken directly orthogonal to the leaves. 
Ground-truth annotations for the number of leaves were automatically determined from the model. Furthermore, these annotations were calibrated to the real images in that it only counted a leaf if it was not occluded (algorithmically determined), and if it was larger than a threshold chosen so that the leaf would be visible by a human. Lastly, only images with at least one leaf were kept, and the resulting dataset is known as the synthetic dataset.  This synthetic dataset contained 1350 images and is referred to as $M^S$. Figure~\ref{fig_met_8} shows a real maize plant image on day 25 from 0 degrees, and a synthetic maize plant image on the same day (Figure~\ref{fig_met_9}).

\paragraph{Canola dataset:} Real images of canola are available at \cite{canolaURL}. The dataset (called P2IRC Flagship 1 Data) contains images of 50 different spring type genotypes of \textit{Brassica napus} (canola), with six replicates of each, grown under two treatment conditions for over 38 days at the LemnaTec Scanalyzer 3D facility at University of Nebraska, Lincoln. Details about the dataset can be found in  \cite{nazifa-jana}. For the inflorescence branch count predictions we selected only top view images and those that had flowers. Also, while this dataset  was a case/control dataset (meaning some plants were watered sufficiently, and some were not to study the affect of drought/stress) the `case'  images had technical issues  \cite{nazifa-jana}. Since the control images were enough for the purposes of this work, we only use the control images here. In total, the real dataset contained 385 images, referred to as $C^R$. The inflorescent branch count annotations were scored manually (details about manual scoring in   \cite{nazifa-jana}). 

\begin{figure}[!htbp]
  \centering
  \subfloat[]{\includegraphics[width=0.3\linewidth, height=0.85in]{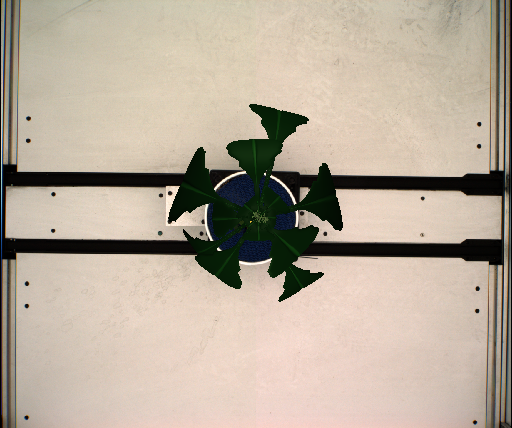}\label{fig_met_1}}
  \hfill
  \subfloat[]{\includegraphics[width=0.3\linewidth, height=0.85in]{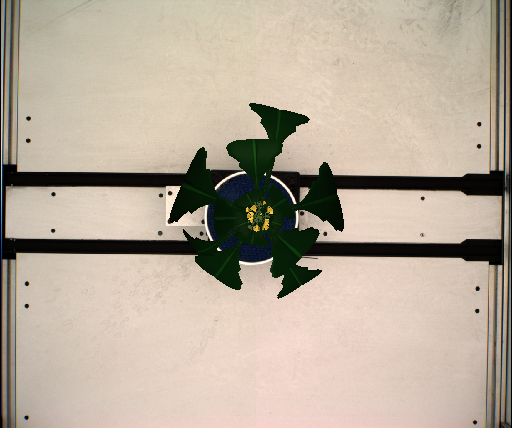}\label{fig_met_2}}
   \hfill
  \subfloat[]{\includegraphics[width=0.3\linewidth, height=0.85in]{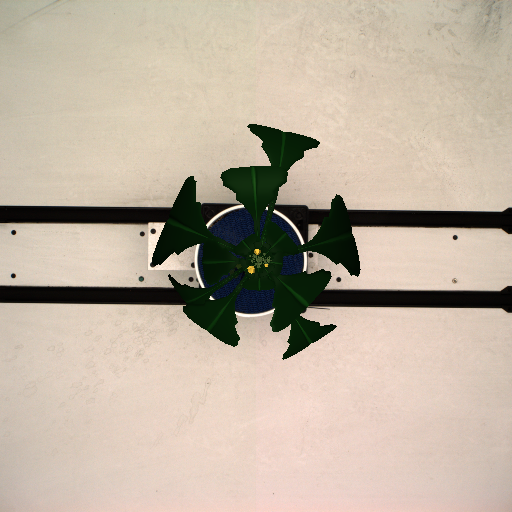}\label{fig_met_3}}
   \hfill
  \subfloat[]{\includegraphics[width=0.3\linewidth, height=0.85in]{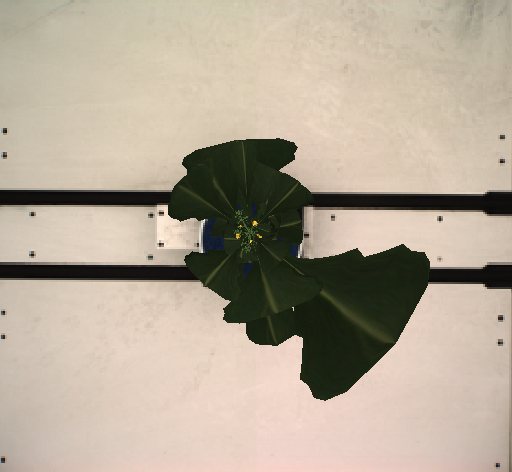}\label{fig_met_4}}
    \hfill
  \subfloat[]{\includegraphics[width=0.3\linewidth, height=0.85in]{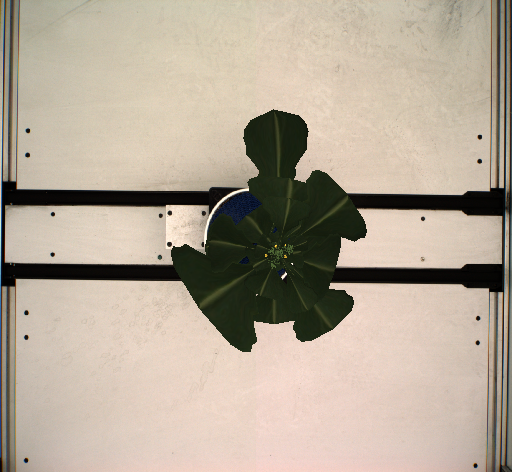}\label{fig_met_5}}
      \hfill
  \subfloat[]{\includegraphics[width=0.3\linewidth, height=0.85in]{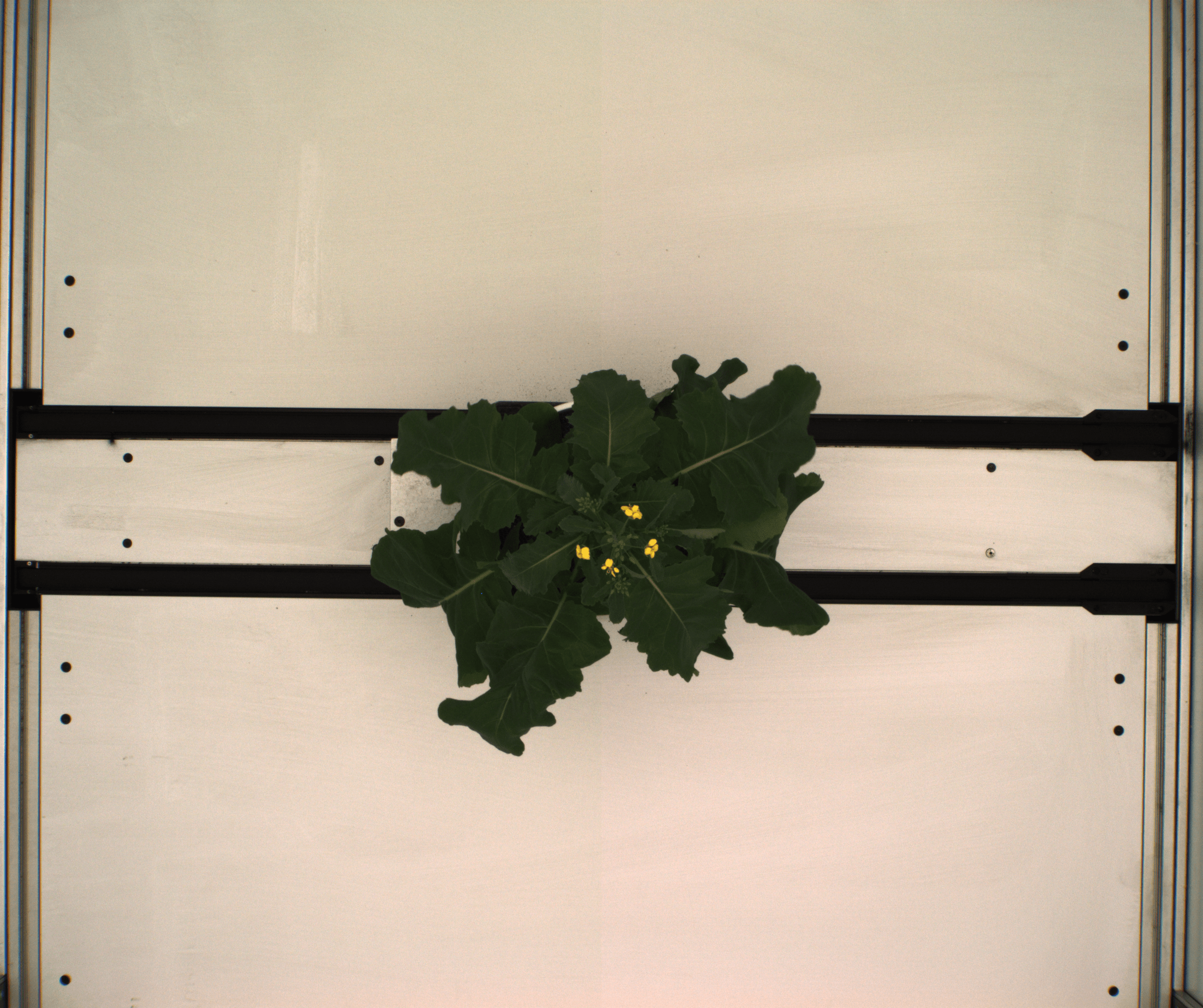}\label{fig_met_6}}
  \caption{(a) -- (e) Synthetic canola images generated from $C^S_1$ through $C^S_5$ respectively, at the same time point. (f) A real canola image at approximately the same time.}
  \label{fig_met_7}
\end{figure}

The development of 200 plants was simulated from the canola L-system in  \cite{mik-nazifa-lsystem} using the same timeline as the real plants. Only flowering images were considered. As canola is a more complex plant to model in terms of leaf architecture and in terms of determining inflorescence branches than in maize, calibrating the model was challenging. Ultimately, we created 5 different synthetic datasets of canola called $C^S_1, C^S_2, C^S_3, C^S_4, C^S_5$, each contained 1200 images generated from five different variants of the canola plant model. These variants were obtained using a refinement procedure as described below. Figure~\ref{fig_met_7} shows five different synthetic images of canola at approximately the same time point together with a real image. This is similar to Klein et al.\ \cite{klein-synthetic} who used synthetic images to train a classifier to distinguish between healthy and infected tomato plants, where they iteratively generated synthetic images to obtain optimal performance. 

\section{Methodology}
\label{sec:method}
\paragraph{Maize study methodology:}
A leaf counting procedure and experiment was performed on maize using the Deep Plant Phenomics (DPP) platform \cite{jordan-ian} using a CNN that was created and trained according to \cite{jordan-tuitorial}. The model structure contained six convolutional layers with filter dimensions $(5,5,3,32)$, $(5,5,32,64)$, $(5,5,32,64)$, $(3,3,64,64)$, $(3,3,64,64)$, and $(3,3,128,128)$, stride length $1$, and the $tanh$ activation function. Each convolutional layer was followed by a pooling layer with kernel size $3$ and stride length $2$. The model parameters and training hyper-parameters were: batch size $4$, image dimensions $256 \times 256$, learning rate $0.0001$, and number of epochs $500$. A data augmentation procedure was done that consisted of cropping, flipping, and adjusting brightness/contrast. The testing network had an additional output layer. We did not further investigate other architectures for better comparison to other works in the literature \cite{jordan,ian-jordan-maize}.

Four sets of experiments were conducted. In the first set, only real images were used for training.  As $M^R$ consisted of 13 plants with one image per day across 27 days, for each $i$ from 1 to 8, training was done by randomly sampling without replacement $i$ of the 13 plants, and then using all of the images of those $i$ plants for training. This was done in contrast to training with a randomly selected subset of images from the entire real image set, as it could be easier for an experimenter to create ground truth annotations for a small number of plants over time than to find and create ground truth annotations for a large number of different plants and randomly selected time points. For each, $5$ of the remaining plants  were used for testing purpose. Testing was done in two different ways. First, testing was performed using 100 randomly selected real images from all of the remaining images of the $5$ plants that were not used for training. Second, testing was performed using all remaining real images. Evaluation was measured by comparing the number of real leaves to the number of predicted leaves using mean absolute loss (mean absolute value of correct number minus predicted number), standard deviation, and the square of Pearson's correlation coefficient ($r^2$). In this second set of experiments, the procedure above was repeated but by augmenting with the synthetic images (but only testing on real images). In the third set, only the synthetic images were used for training but using real images for testing. The fourth experiment used 8 synthetic plants for training with testing on synthetic images. 

\paragraph{Canola study methodology:}
A similar experiment was performed on canola using inflorescence branch count. The CNN model and parameters were the same as for maize. As there are 39 distinct genotypes, we use images from $i$ genotypes, in increments of 3 for training (again, using all images from each selected genotype). For example, the first training set has images of flowering plants from $3$ genotypes, and the next training set has images of flowering from $6$ genotypes, and so on. Testing was done in two different ways. Testing was performed using $100$ randomly selected real images from all images not used for training, and also using all remaining real images. Evaluation was again done between number of real and predicted inflorescent branch numbers using mean absolute loss, standard deviation, and $r^2$. This entire procedure was repeated five additional times by using each of $C^S_1, C^S_2, C^S_3, C^S_4, C^S_5$ respectively.  

\section{Results}
\label{sec:result}

\paragraph{Maize study results:}

The results regarding prediction of  maize leaf count are in Table~\ref{tab_1}.  Of special note is the row with $0$ plants, as that corresponds to only training on synthetic data but testing on real data.

\begin{table*}[htbp!]
\centering
\resizebox{\textwidth}{!}{%
\begin{tabular}{|c|cccc|}
\hline
{\color[HTML]{000000} }                                                                                                       & \multicolumn{4}{c|}{{\color[HTML]{000000} Mean Absolute Loss (Absolute Loss Standard Deviation, $r^2$)}}                                                                                                                                                                                                                                                                                                                                                                           \\ \cline{2-5} 
{\color[HTML]{000000} }                                                                                                       & \multicolumn{2}{c|}{{\color[HTML]{000000} Tested with 100 real images}}                                                                                                                                                                            & \multicolumn{2}{c|}{{\color[HTML]{000000} Tested with remaining real images}}                                                                                                                                                 \\ \cline{2-5} 
\multirow{-3}{*}{{\color[HTML]{000000} \begin{tabular}[c]{@{}c@{}}Numbers of real plants\\used in Training\\(number of real images)\end{tabular}}} & \multicolumn{1}{c|}{\begin{tabular}[c]{@{}c@{}}No synthetic images\\used for training\end{tabular}} & \multicolumn{1}{c|}{\begin{tabular}[c]{@{}c@{}}All synthetic images\\used for training\end{tabular}} & \multicolumn{1}{c|}{\begin{tabular}[c]{@{}c@{}}No synthetic images\\used for training\end{tabular}} & \begin{tabular}[c]{@{}c@{}}Synthetic images\\used for training\end{tabular} \\ \hline
{\color[HTML]{000000} 0 real plant}                                                                                           & \multicolumn{1}{c|}{{\color[HTML]{000000} --}}                                                                          & \multicolumn{1}{c|}{{\color[HTML]{000000} 1.02 (0.74, 0.62)}}                                                   & \multicolumn{1}{c|}{{\color[HTML]{000000} --}}                                                                          & {\color[HTML]{000000} 0.76 (0.69, 0.78)}                                                   \\ \hline
{\color[HTML]{000000} 1 real plant (26)}                                                                                           & \multicolumn{1}{c|}{{\color[HTML]{000000} 1.33 (0.84, 0.41)}}                                                  & \multicolumn{1}{c|}{{\color[HTML]{000000} 0.64 (0.68, 0.79)}}                                                   & \multicolumn{1}{c|}{{\color[HTML]{000000} 1.20 (0.80, 0.57)}}                                                  & {\color[HTML]{000000} 0.51 (0.58, 0.87)}                                                   \\ \hline
{\color[HTML]{000000} 2 real plant (51)}                                                                                           & \multicolumn{1}{c|}{{\color[HTML]{000000} 1.03 (0.69, 0.63)}}                                                           & \multicolumn{1}{c|}{{\color[HTML]{000000} 0.58 (0.61, 0.83)}}                                                            & \multicolumn{1}{c|}{{\color[HTML]{000000} 1.01 (0.71, 0.67)}}                                                           & {\color[HTML]{000000} 0.55 (0.59, 0.85)}                                                            \\ \hline
{\color[HTML]{000000} 3 real plant (77)}                                                                                           & \multicolumn{1}{c|}{{\color[HTML]{000000} 0.95 (0.73, 0.65)}}                                                           & \multicolumn{1}{c|}{{\color[HTML]{000000} 0.67 (0.67, 0.78)}}                                                            & \multicolumn{1}{c|}{{\color[HTML]{000000} 0.75 (0.70, 0.75)}}                                                           & {\color[HTML]{000000} 0.57 (0.62, 0.83)}                                                            \\ \hline
{\color[HTML]{000000} 4 real plant (103)}                                                                                           & \multicolumn{1}{c|}{{\color[HTML]{000000} 0.67 (0.61, 0.80)}}                                                           & \multicolumn{1}{c|}{{\color[HTML]{000000} 0.55 (0.62, 0.83)}}                                                            & \multicolumn{1}{c|}{{\color[HTML]{000000} 0.60 (0.63, 0.83)}}                                                           & {\color[HTML]{000000} 0.50 (0.60, 0.86)}                                                            \\ \hline
{\color[HTML]{000000} 5 real plant (126)}                                                                                           & \multicolumn{1}{c|}{{\color[HTML]{000000} 0.86 (0.74, 0.69)}}                                                           & \multicolumn{1}{c|}{{\color[HTML]{000000} 0.54 (0.66, 0.82)}}                                                            & \multicolumn{1}{c|}{{\color[HTML]{000000} 0.73 (0.71, 0.77)}}                                                           & {\color[HTML]{000000} 0.44 (0.58, 0.88)}                                                            \\ \hline
{\color[HTML]{000000} 6 real plant (150)}                                                                                           & \multicolumn{1}{c|}{{\color[HTML]{000000} 0.64(0.64, 0.79)}}                                                            & \multicolumn{1}{c|}{{\color[HTML]{000000} 0.59 (0.64, 0.81)}}                                                            & \multicolumn{1}{c|}{{\color[HTML]{000000} 0.57 (0.65, 0.84)}}                                                           & {\color[HTML]{000000} 0.47 (0.60, 0.87)}                                                            \\ \hline
{\color[HTML]{000000} 7 real plant (175)}                                                                                           & \multicolumn{1}{c|}{{\color[HTML]{000000} 0.58 (0.68, 0.81)}}                                                  & \multicolumn{1}{c|}{{\color[HTML]{000000} 0.53 (0.65, 0.83)}}                                                   & \multicolumn{1}{c|}{{\color[HTML]{000000} 0.56 (0.63, 0.84)}}                                                  & {\color[HTML]{000000} 0.46 (0.61, 0.87)}                                                            \\ \hline
{\color[HTML]{000000} 8 real plant (201)}                                                                                           & \multicolumn{1}{c|}{{\color[HTML]{000000} 0.48 (0.55, 0.87)}}                                                           & \multicolumn{1}{c|}{{\color[HTML]{000000} 0.42 (0.53, 0.89)}}                                                            & \multicolumn{1}{c|}{{\color[HTML]{000000} 0.50 (0.57, 0.86)}}                                                           & {\color[HTML]{000000} 0.45 (0.54, 0.88)}                                                            \\ \hline
\end{tabular}%
}
\caption{The mean absolute loss (respectively standard deviation, $r^2$) of leaf number prediction in maize appears in each cell. The first column indexes the number of real plants (number of images in parentheses) used for training. Columns 2 and 4 are for using no additional synthetic data for training, while columns 3 and 5 augment the amount of indexed real data by all the synthetic data. Results when testing with 100 real images (respectively all remaining images) are in the second and third column (respectively fourth and fifth column).}
\label{tab_1}
\end{table*}

\begin{figure}[!htbp]
\center
  \includegraphics[width=1\linewidth]{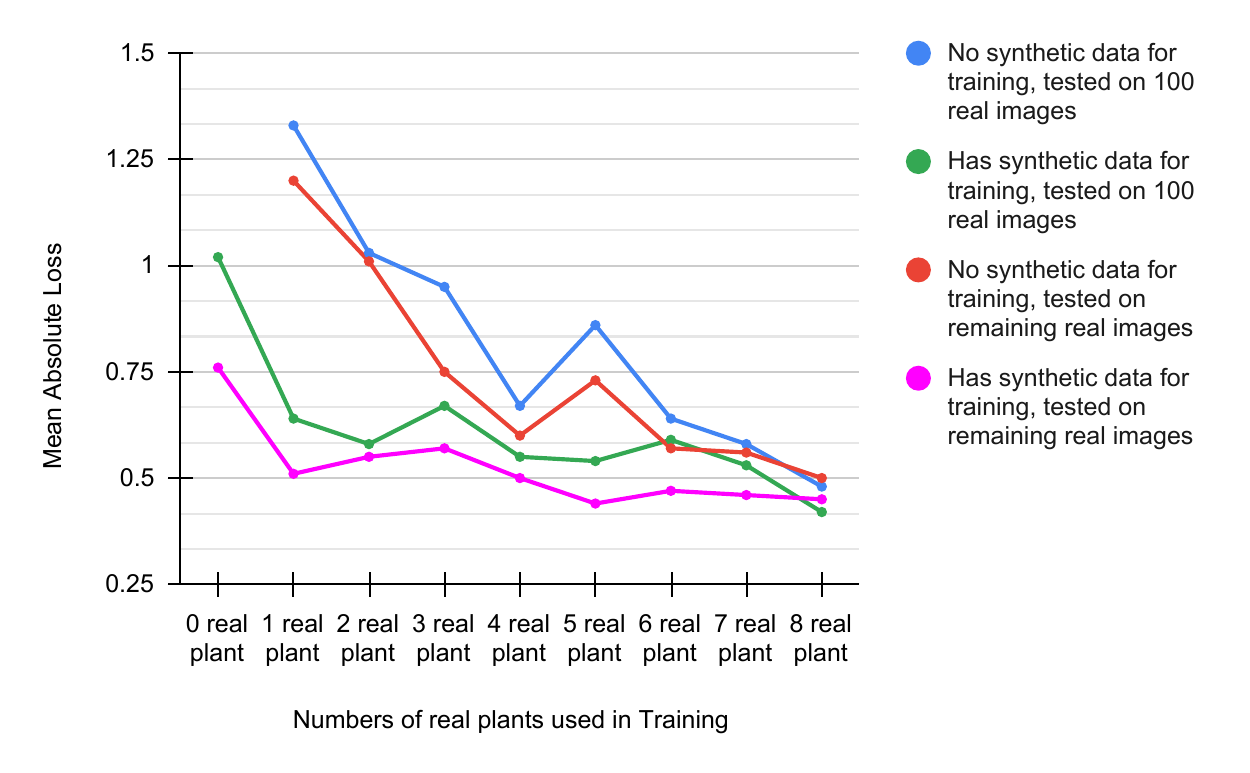}
  \caption{The mean absolute losses for leaf counting in maize from Table~\ref{tab_1} when training with only real images and testing with 100 real images (blue color), training with synthetic images plus (optionally) some number of real images and testing with 100 real images (green color), training with only real images and testing with remaining real images (red color), and training with synthetic images plus (optionally) some number of real images and testing with remaining real images (pink color).}
  \label{fig_12}
\end{figure}

Figure~\ref{fig_12} contrasts the mean absolute losses described in Table~\ref{tab_1}. The blue and green graphs show the mean absolute loss obtained on testing with 100 real images and training with real images only (blue), and training with  synthetic images plus (optionally) some number of real images. The red and pink graphs of the figure shows a similar contrast while testing with all remaining images, where the red graph is for no synthetic images augmentation. Another experiment was done training the network with 8 synthetic plants only and no real images. Then, the trained model was tested with 100 synthetic images, and with all remaining synthetic images. The obtained results are $0.08\ (0.27, 0.98)$, and $0.05\ (0.23, 0.99)$ for mean absolute loss (respectively standard deviation, $r^2$). Finally, we calculated the Root Mean Squared Error (RMSE) while training with 8 real maize plants combined with all  synthetic data, and tested on 100 real maize images, and obtained the value of $1.16$. 

\paragraph{Canola study results:}
The inflorescence branch count predictions for canola were calculated. Table~\ref{tab_2} shows the results when testing on 100 real images only. The results when testing on all remaining images are quite similar and appear in Supp Table~\ref{tab_3}. Here, the results are shown from each of $C^S_1, C^S_2, C^S_3, C^S_4, C^S_5$. The first L-system variant was created first. After seeing the results and visually comparing the real and synthetic image differences, this was used to create model variant 2, etc, through variant 5. The positions where augmenting with synthetic data at least matches real data only are bolded in Table~\ref{tab_2}.

\begin{table*}[!htbp]
\centering
\resizebox{\textwidth}{!}{%
\begin{tabular}{|c|cccccc|}
\hline
\multirow{2}{*}{\begin{tabular}[c]{@{}c@{}}Numbers of real plant genotypes\\ (Number of real plants,\\ Number of real images) \\ used in Training\end{tabular}} & \multicolumn{6}{c|}{Mean Absolute Loss (Absolute Loss Standard Deviation, $r^2$), tested on 100 real images}                                                                                                                                                                                                                                                                                                                                                                                                                                                                                                                                                                                                       \\ \cline{2-7} 
                                                                                                                                                                & \multicolumn{1}{c|}{\begin{tabular}[c]{@{}c@{}}No synthetic data\\ augmented for \\ training\end{tabular}} & \multicolumn{1}{c|}{\begin{tabular}[c]{@{}c@{}}Synthetic data\\$C^S_1$ augmented\\for training\end{tabular}} & \multicolumn{1}{c|}{\begin{tabular}[c]{@{}c@{}}Synthetic data\\$C^S_2$ augmented\\for training\end{tabular}} & \multicolumn{1}{c|}{\begin{tabular}[c]{@{}c@{}}Synthetic data\\$C^S_3$ augmented\\for training\end{tabular}} & \multicolumn{1}{c|}{\begin{tabular}[c]{@{}c@{}}Synthetic data\\$C^S_4$ augmented\\for training\end{tabular}} & \begin{tabular}[c]{@{}c@{}}Synthetic data\\$C^S_5$ augmented\\for training\end{tabular} \\ \hline
0 real data                                                                                                                                                     & \multicolumn{1}{c|}{--}                                                                                    & \multicolumn{1}{c|}{12.48 (8.78, -7.22)}                                                                               & \multicolumn{1}{c|}{16.61 (5.63, -9.86)}                                                                               & \multicolumn{1}{c|}{1.98 (2.25, 0.68)}                                                                                  & \multicolumn{1}{c|}{\textbf{1.45 (1.76, 0.81)}}                                                                        & 5.6 (3.63, -0.57)                                                                                \\ \hline
3 genotypes (9, 32)                                                                                                                                             & \multicolumn{1}{c|}{1.27 (1.51, 0.86)}                                                            & \multicolumn{1}{c|}{5.34 (6.77, -1.62)}                                                                                & \multicolumn{1}{c|}{2.81 (3.55, 0.27)}                                                                                 & \multicolumn{1}{c|}{1.73 (2.02, 0.74)}                                                                                  & \multicolumn{1}{c|}{1.88 (2.17, 0.70)}                                                                                 & 2.31 (2.62, 0.56)                                                                                \\ \hline
6 genotypes (17, 63)                                                                                                                                            & \multicolumn{1}{c|}{1.1 (1.22, 0.9)}                                                                       & \multicolumn{1}{c|}{3.15 (4.15, 0.04)}                                                                                 & \multicolumn{1}{c|}{3.38 (4.05, 0.02)}                                                                                 & \multicolumn{1}{c|}{1.33 (1.60, 0.84)}                                                                                  & \multicolumn{1}{c|}{1.42 (1.58, 0.83)}                                                                                 & 1.76 (2.24, 0.71)                                                                                \\ \hline
9 genotypes (26, 80)                                                                                                                                            & \multicolumn{1}{c|}{0.97 (0.95, 0.93)}                                                                     & \multicolumn{1}{c|}{2.21 (3.17, 0.47)}                                                                                 & \multicolumn{1}{c|}{1.61 (2.11, 0.75)}                                                                                 & \multicolumn{1}{c|}{1.18 (1.49, 0.87)}                                                                                  & \multicolumn{1}{c|}{1.29 (1.76, 0.83)}                                                                                 & 1.61 (2.24, 0.72)                                                                                \\ \hline
12 genotypes (34, 101)                                                                                                                                          & \multicolumn{1}{c|}{0.8 (1.01, 0.94)}                                                                      & \multicolumn{1}{c|}{1.65 (2.73, 0.64)}                                                                                 & \multicolumn{1}{c|}{1.36 (2.32, 0.74)}                                                                                 & \multicolumn{1}{c|}{1.16 (1.5, 0.87)}                                                                                   & \multicolumn{1}{c|}{1.23 (1.57, 0.86)}                                                                                 & 1.26 (1.93, 0.81)                                                                                \\ \hline
15 genotypes (43, 123)                                                                                                                                          & \multicolumn{1}{c|}{0.95 (1.13, 0.92)}                                                                     & \multicolumn{1}{c|}{1.28 (2.33, 0.75)}                                                                                 & \multicolumn{1}{c|}{1.42 (1.81, 0.81)}                                                                                 & \multicolumn{1}{c|}{0.92 (1.19, 0.91)}                                                                                  & \multicolumn{1}{c|}{\textbf{0.89 (1.37, 0.90)}}                                                                                 & 1.56 (2.35, 0.71)                                                                                \\ \hline
18 genotypes (50, 146)                                                                                                                                          & \multicolumn{1}{c|}{0.84 (1.03, 0.93)}                                                                     & \multicolumn{1}{c|}{1.2 (2.23, 0.77)}                                                                                  & \multicolumn{1}{c|}{1.23 (1.62, 0.85)}                                                                                 & \multicolumn{1}{c|}{0.99 (1.34, 0.90)}                                                                                  & \multicolumn{1}{c|}{1.08 (1.38, 0.89)}                                                                                 & 1.25 (1.94, 0.81)                                                                                \\ \hline
21 genotypes (59, 165)                                                                                                                                          & \multicolumn{1}{c|}{0.69 (0.78, 0.96)}                                                                     & \multicolumn{1}{c|}{1.58 (2.19, 0.74)}                                                                                 & \multicolumn{1}{c|}{1.42 (1.83, 0.81)}                                                                                 & \multicolumn{1}{c|}{0.9 (1.39, 0.9)}                                                                                    & \multicolumn{1}{c|}{0.91 (1.37, 0.90)}                                                                                 & 1.04 (1.83, 0.84)                                                                                \\ \hline
24 genotypes (66, 184)                                                                                                                                          & \multicolumn{1}{c|}{0.83 (1.17, 0.92)}                                                                     & \multicolumn{1}{c|}{1.52 (2.18, 0.74)}                                                                                 & \multicolumn{1}{c|}{1.12 (1.47, 0.87)}                                                                                 & \multicolumn{1}{c|}{0.91 (1.32, 0.90)}                                                                                  & \multicolumn{1}{c|}{0.93 (1.42, 0.89)}                                                                                 & 1.04 (2.03, 0.81)                                                                                \\ \hline
27 genotypes (73, 200)                                                                                                                                          & \multicolumn{1}{c|}{0.78 (1.05, 0.93)}                                                                     & \multicolumn{1}{c|}{1.15 (1.94, 0.82)}                                                                                 & \multicolumn{1}{c|}{0.91 (1.49, 0.89)}                                                                                 & \multicolumn{1}{c|}{0.96 (1.57, 0.87)}                                                                                  & \multicolumn{1}{c|}{\textbf{0.77 (1.05, 0.93)}}                                                                                 & 1.08 (1.70, 0.85)                                                                                \\ \hline
30 genotypes (80, 215)                                                                                                                                          & \multicolumn{1}{c|}{0.63 (0.83, 0.96)}                                                                     & \multicolumn{1}{c|}{0.99 (2.02, 0.82)}                                                                                 & \multicolumn{1}{c|}{0.88 (1.65, 0.87)}                                                                                 & \multicolumn{1}{c|}{0.95 (1.43, 0.89)}                                                                                  & \multicolumn{1}{c|}{0.89 (1.20, 0.92)}                                                                                 & 0.98 (1.84, 0.84)                                                                                \\ \hline
33 genotypes (89, 236)                                                                                                                                          & \multicolumn{1}{c|}{0.66 (0.96, 0.95)}                                                                     & \multicolumn{1}{c|}{1.23 (1.91, 0.81)}                                                                                 & \multicolumn{1}{c|}{0.82 (1.21, 0.92)}                                                                                 & \multicolumn{1}{c|}{0.92 (1.27, 0.91)}                                                                                  & \multicolumn{1}{c|}{0.82 (1.09, 0.93)}                                                                                 & 0.85 (1.60, 0.88)                                                                                \\ \hline
36 genotypes (96, 254)                                                                                                                                          & \multicolumn{1}{c|}{0.84 (0.97, 0.94)}                                                                     & \multicolumn{1}{c|}{1.12 (1.99, 0.81)}                                                                                 & \multicolumn{1}{c|}{0.76 (1.09, 0.93)}                                                                                 & \multicolumn{1}{c|}{1.11 (1.61, 0.86)}                                                                                  & \multicolumn{1}{c|}{\textbf{0.84 (1.26, 0.91)}}                                                                                 & 1.02 (1.66, 0.86)                                                                                \\ \hline
39 genotypes (104, 285)                                                                                                                                         & \multicolumn{1}{c|}{0.7 (0.87, 0.95)}                                                             & \multicolumn{1}{c|}{1.04 (2.17, 0.79)}                                                                                 & \multicolumn{1}{c|}{0.9 (1.26, 0.91)}                                                                                  & \multicolumn{1}{c|}{0.99 (1.31, 0.9)}                                                                                   & \multicolumn{1}{c|}{0.75 (1.16, 0.93)}                                                                        & 0.92 (1.71, 0.86)                                                                                \\ \hline
\end{tabular}%
}
\caption{Each table cell contains the mean absolute loss (respectively standard deviation and $r^2$) for the inflorescence branch counting task in canola. Training was done on real images as indexed in the first column, column 2 is where it was not augmented by any synthetic data, and column 3 through 7 are augmented by synthetic data from  $C^S_1, C^S_2, C^S_3, C^S_4, C^S_5$ respectively.}
\label{tab_2}
\end{table*}

\begin{figure}[!htbp]
\center
  \includegraphics[width=\linewidth]{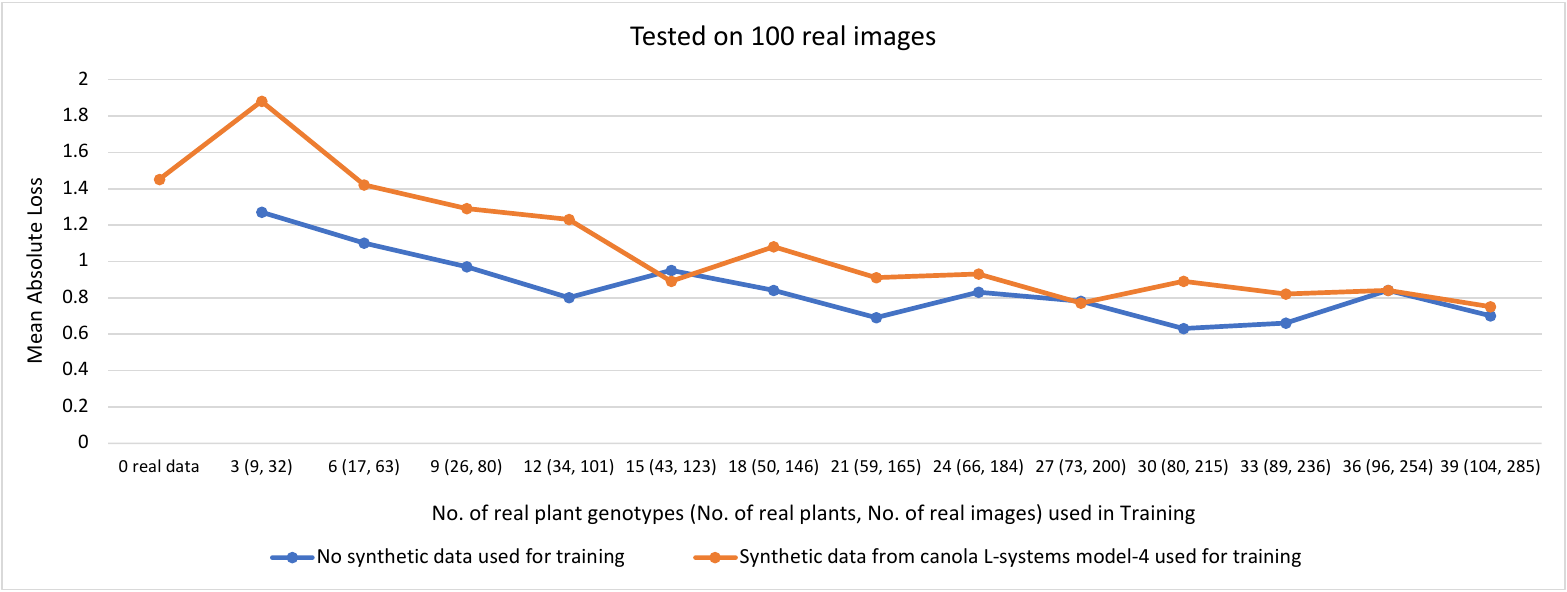}
  \caption{A contrast of the mean absolute losses in Table~\ref{tab_2} (second column  and sixth column) while training with only real images and testing with 100 images (blue color) with the mean absolute loss while training with synthetic images from $C^S_4$ plus (optionally) some number of real images (orange color) for the inflorescence branch counting task in canola.}
  \label{fig_13}
\end{figure}

Figure~\ref{fig_13} contrasts the mean absolute losses described in second and sixth column of Table~\ref{tab_2}. The blue and orange graphs show the mean absolute loss obtained on testing with 100 real images and training with real images only (blue), and training with synthetic images $C^S_4$ generated from fourth model plus (optionally) some number of real images.

Another experiment was done where the neural network was trained on 960 ($80\%$) of the synthetic images in $C^S_4$, and then tested on the other 240 ($20\%$) of the synthetic images in $C^S_4$, obtaining $1.19\ (1.26, 0.94)$.

\subsection{Refinement of Canola L-systems}
\begin{figure}[!htbp]
  \centering
  \subfloat[]{\includegraphics[width=0.3\linewidth, height=0.85in]{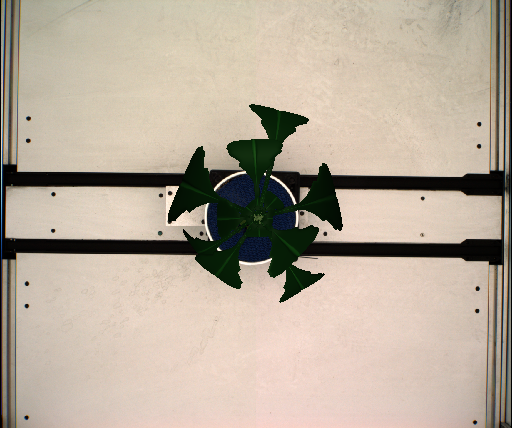}\label{fig_met_14}}
  \hfill
  \subfloat[]{\includegraphics[width=0.3\linewidth, height=0.85in]{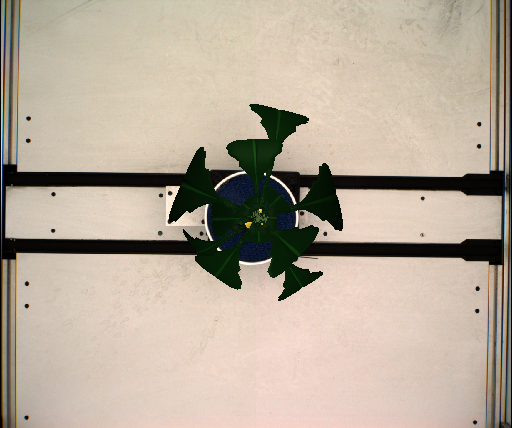}\label{fig_met_15}}
   \hfill
  \subfloat[]{\includegraphics[width=0.3\linewidth, height=0.85in]{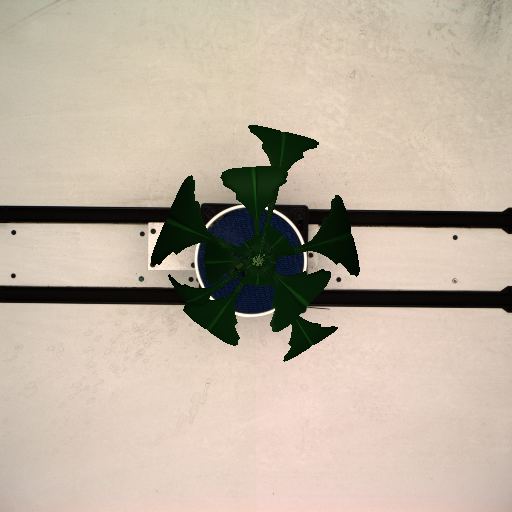}\label{fig_met_16}}
   \hfill
  \subfloat[]{\includegraphics[width=0.3\linewidth, height=0.85in]{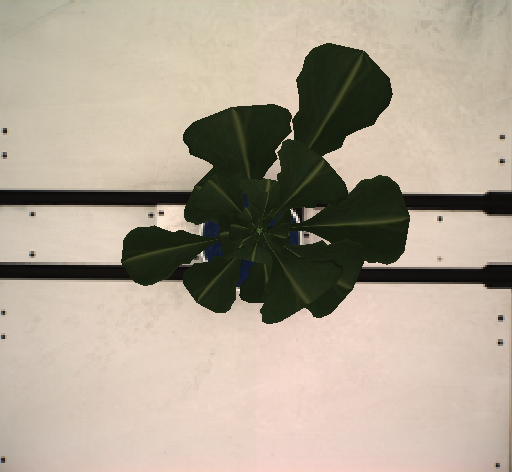}\label{fig_met_17}}
    \hfill
  \subfloat[]{\includegraphics[width=0.3\linewidth, height=0.85in]{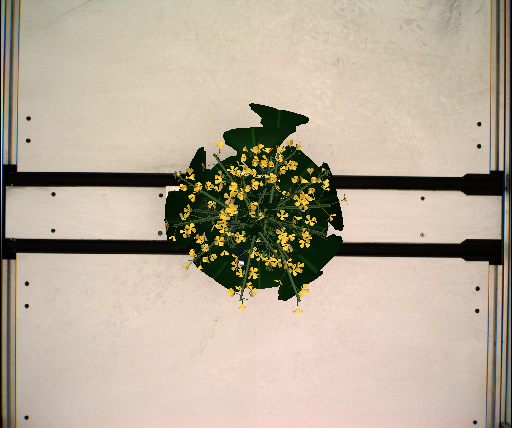}\label{fig_met_18}}
      \hfill
  \subfloat[]{\includegraphics[width=0.3\linewidth, height=0.85in]{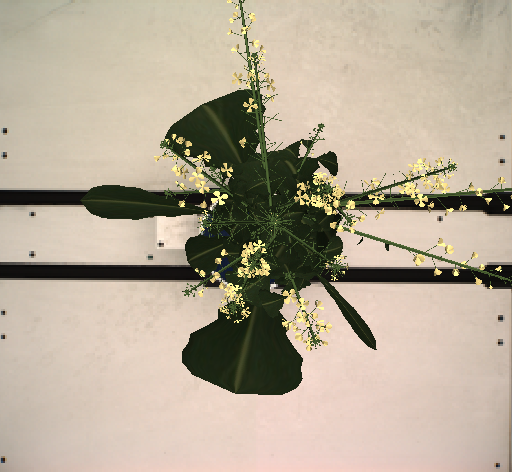}\label{fig_met_19}}
  \caption{Synthetic images of canola on the same day generated from four different procedural models of canola plants (a -- d) and two flowering images of same time point from different models. (a) $C^S_1$ (b) $C^S_2$ (c) $C^S_3$ (d) $C^S_4$ (e) $C^S_1$ (f) $C^S_4$.}
  \label{fig_met_20}
\end{figure}

Feedback from the poor prediction results in earlier L-system variants and from visual inspection between real and synthetic images of canola helped to create refined L-system variants. For example, Figure~\ref{fig_met_20} shows some synthetic images of canola from the same day generated from four different variants (created by recalibration) of same procedural model of canola plants (Figure~\ref{fig_met_14} from $C^S_1$, \ref{fig_met_15} from $C^S_2$, \ref{fig_met_16} from $C^S_3$, and \ref{fig_met_17} from $C^S_4$). A closer look at these figures shows how leaf shape and curvature were adjusted using L-system parameters to generate more realistic looking canola leaves, which substantially helped in improving the deep learning results. Another scenario is shown in  Figure~\ref{fig_met_18} and  \ref{fig_met_19}, where the inflorescence branches and flowers were improved from the first to the fourth model. Supp Figure~\ref{fig_23} shows how the distribution of inflorescence branch numbers changed from the first to through the fourth model.

Specifically, the first L-system was from \cite{mik-nazifa-lsystem} (only modified to count  inflorescence branch numbers). The second was created by changing the {\it branch vigour} parameter with the mean value and standard deviation being increased to obtain greater variability of the feature; this changed plant morphology. From the second to third L-system, the canola model parameters were changed to generate a distribution of inflorescence branch numbers to better reflect the real dataset. Supp Figure~\ref{fig_24} shows the distribution of inflorescence branch counts in the real canola images. If we look at Supp Figures~\ref{fig_23} and \ref{fig_24}, we can see how the inflorescence branch number distribution was adjusted from first to the fourth L-system according to the real data distribution. Additionally, the mean value and standard deviation were made slightly larger to decrease the number of lateral branches, and to have better control over the apex behaviour. From the third to fourth L-system, there were a number of adjustments. The means and standard deviations of growth, branching, and leaf parameters were changed. The user-defined functions for growth of all organs (e.g., leaves and internodes) were stretched so organs grow over a longer period. The function used to determine leaf width was changed. The leaf texture was also changed from a simple green colour with a white midvein to an image-based colour (i.e., the leaf texture was extracted from a real canola image). Hence, the fourth model was made substantially more realistic and was more accurately calibrated to the real images, compared to the first or second model. For the last L-system, few changes were made, such as to petal color, and leaf size differences.

\section{Discussion}
In maize, by comparing the model's mean absolute loss for real and synthetic datasets (blue and green lines) in Figure~\ref{fig_12} (as with the red and pink lines), it is evident that for a fixed number of real plants, augmenting with synthetic images always improved results. This was especially true when there were fewer real plants, such as 2 (51 images) where adding synthetic images cut the mean absolute loss almost in half from $1.01$ to $0.55$. Training with only synthetic images requires no groundtruth labelling, and provided a better mean absolute loss than training with 51 images. Also, training with only one real plant (26 images) and synthetic images had a mean absolute loss of $0.51$ which was comparable to training with eight real plants (201 images) with a loss of $0.45$. Hence, augmenting with synthetic images can significantly reduce the amount of labelled real images needed, and they can be outright eliminated in some cases without a large cost in accuracy. The results also present an overview of the minimum number of real plant images needed to get a desired result.

\begin{figure}[!htbp]
  \center
  \includegraphics[width=0.95\linewidth]{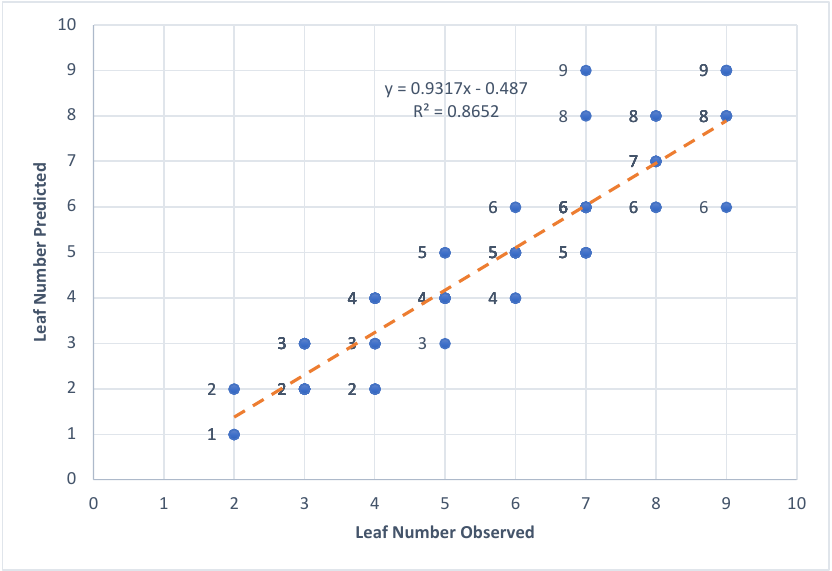} 
  \caption{Observed (along X axis) and predicted (along Y axis) leaf count in maize when tested on 100 real maize images, and trained on 8 real maize plants plus synthetic maize images. The numbers beside the data points show the frequency of that data points.}
  \label{fig_21}
\end{figure}

It should be noted that it was relatively easy to produce visually realistic synthetic images of maize due to its simpler architecture, and indeed using these synthetic images was quite helpful. We achieved a RMSE of $1.16$. In contrast,  in \cite{ian-jordan-maize} the RMSE obtained using their synthetic maize images was between $1.50$ to $1.75$ with almost equal amounts of real data augmented with synthetic images. Their $r^2$ ranged from between 0.5 and 0.75, whereas ours mostly ranged from 0.62 and 0.89 after augmentation. The predictions in \cite{ian-jordan-maize} were sometimes worse when augmenting with synthetic images, and our results were always improved, and our RMSE and $r^2$ were better. While their real images of maize were taken at the same imaging facility, we caution against any firm conclusions as their real images were different and include up until 66 days after planting, versus 29 days in our dataset  which could affect predictions. However, their often worse prediction results when augmenting with synthetic images were likely partially caused by less realistic synthetic images. Indeed, they indicated that despite maize following alternating phyllotaxy, their procedurally generated images had successive leaves that emerged from the stalk at random angles \cite{ian-jordan-maize}. Our L-system was built to properly follow alternating phyllotaxy. This shows the importance of biological realism in creating synthetic images. Figure~\ref{fig_21} plots real leaf count versus predicted count, and shows the frequency of the data points. 

\begin{figure}[!htbp]
  \center
  \includegraphics[width=0.9\linewidth]{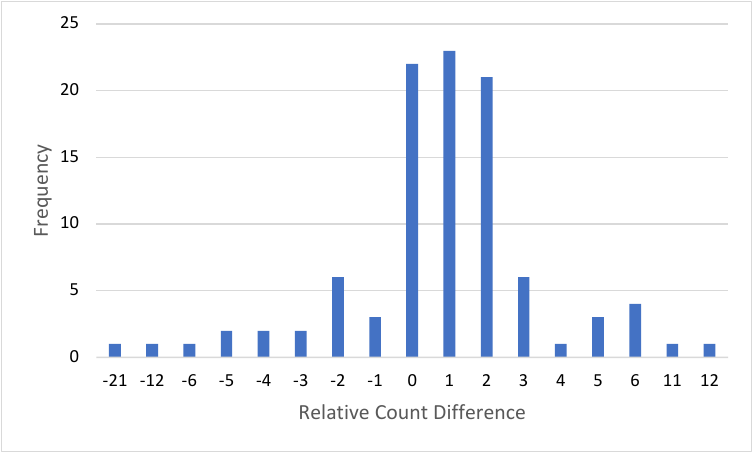} 
  \caption{Distribution of relative count difference (correct number - predicted number) for inflorescence branch count in canola, tested on 100 real canola images, after training with all the remaining real images (285) augmented with $C^S_4$. This result is from the last row and sixth column of Table~\ref{tab_2}.}
  \label{fig_22}
\end{figure}

For canola, results were mixed. There were later stage images with more overlapping and occluded components with flowering images.  In addition, canola has a more a complex branching pattern, so the CNN has to distinguish different orders of branches.  The fourth L-system variant showed by far the most promising results. While there were some cases where adding synthetic data led to a slightly better prediction, it mostly was not helping for all five model variants constructed. There are however several points of interest. Figure~\ref{fig_22} presents the distribution of relative count difference for inflorescence branch count in canola, while testing on 100 real canola images and trained on all the remaining real images augmented with $C^S_4$. 

The columns of Table~\ref{tab_2} demonstrates the improvements made to the L-systems, thereby leading to improved deep learning results. Therefore, this exercise of obtaining deep learning results followed by a manual visual inspection and calibration of the model had a great role in improving the L-system itself and further results. Pearson's $r^2$ values show that the synthetic images from the fourth canola L-system were quite similar to the real canola images as the Pearson's correlation coefficients indicates the strength of associations between variables. The $r^2 = 0.86$ when trained on 32 real images and tested on 100 real images, and $ r^2 = 0.81$ when trained on the $1200$ synthetic images generated from $C^S_4$ and tested on the same 100 real images. 

It is evident that these results were quite sensitive to the precise L-system construction. For example, when trained on only $C^S_2$, the mean absolute loss was exceptionally large at $16.6$, but it was quite good when trained only on $C^S_4$ at $1.45$. This again shows that producing realistic synthetic images is extremely important. It also shows the simulated images alone even in canola can perform quite well, while requiring no groundtruth labelling.

One interesting contrast between the maize and canola experiments is that when both training and testing on only synthetic images, the mean absolute loss in maize was extremely low at $0.08$, whereas in canola with $C^S_4$, it was much higher at $1.19$. This seems to indicate that the canola L-system is considerably more complicated and more difficult for the ANN to make accurate predictions. This was one of the important points of discussion in \cite{jordan} regarding the \textit{A. thaliana} model.

Finally, we chose to always train with all the available synthetic images when training rather than smaller numbers (such as the same number of real images). This was done because it is easy to generate arbitrarily  large sets of synthetic images from the same L-system. Also, by contrasting Table~\ref{tab_2} and Supp Table~\ref{tab_4}, the latter being a test to see if the amount of synthetic images played a role in prediction accuracy, showed that it was better to always augment by all the synthetic images rather than gradually adding them in.

\section{Conclusion}
Machine learning and L-systems plant modelling are two broad and impactful fields in plant phenotyping research. Here,  we  showed how  each can help the other. This study conducted computational experiments in maize and canola regarding the use of synthetic images of plants for training artificial neural networks (ANNs). For the task of leaf counting in maize, we calculated the mean absolute value of the difference between the predicted and correct number of leaves (mean absolute loss). We found that  augmenting with synthetic images was always beneficial, and it was especially so if there were fewer real images, with the benefit gradually shrinking as the number of real images increased. This confirms similar findings from \cite{nikolenko2021synthetic} but in the plant phenotyping domain. The L-system used to create synthetic images was quite realistic, following the real architecture of maize, and this realism may have been a key contributor to its improved success versus others in the literature. This is similar to van Breugel et al.\ who showed \cite{van2023synthetic} how imperfection in synthetic data affects  downstream machine language tasks. Furthermore, it was more accurate to only use synthetic images with no groundtruth labelling than using 51 real images. This confirms the potential benefits for using synthetic images for training ANNs. 

The task of inflorescence branch counting in canola was conducted and results were evaluated using a similar procedure. This study again showed the importance of realism. In some cases, the L-system models require calibration, and results can be very  sensitive to changes in the models. Small changes in L-systems can improve the synthetic images remarkably. Here, we also saw that the deep learning results served to quantify how replaceable the synthetic images were for real images rather than using only visual inspections. For example, synthetic images alone from an early L-system variant when tested on real images resulted in a high mean absolute loss of $16.61$. After refinement of the L-system to make the model calibration more accurate, and the images more realistic, this number was reduced to $1.45$.  While augmenting with synthetic data only sometimes improved results, this result also shows that using only synthetic images worked reasonably well at predicting on real image, while requiring no groundtruth labelling.

\section{Future Work}
This study opens the door for many potential future research directions. Directly, this includes more refinement to the canola L-systems. Secondly, we would like to further test the level of realism required for synthetic images to become useful for training purposes in additional dimensions beyond those explored between the five L-systems variants.  Another interesting objective would be to study advantages of it being comparatively easy to adjust an L-system to a new environment \cite{mik-nazifa-lsystem}. One could compare an ANN trained on real images from a different environment from the test set to those trained with synthetic images  from the same environment as the test set. If it is better to use synthetic images from the correct environment, then this provides an opportunity to only use one L-system-based model while only needing to adjust the environment for every new use of phenotyping, instead of manually labelling a new large set of images with the new environment.


\vfil\eject

{
    \small
    \begingroup
    \raggedright
    \bibliographystyle{ieeenat_fullname}
    \bibliography{plant}
    \endgroup
}

\input{X_suppl}

\end{document}

%% file: preamble.tex
\usepackage[dvipsnames]{xcolor}


%% file: X_suppl.tex
\clearpage
\setcounter{page}{1}
\maketitlesupplementary

\begin{table*}[]
\setcounter{table}{0}
\centering
\resizebox{\columnwidth}{!}{%
\begin{tabular}{|c|cc|}
\hline
\multirow{2}{*}{\begin{tabular}[c]{@{}c@{}}Numbers of real plant genotypes \\ (Number of real plants, Number \\ of real images) + Numbers of \\ synthetic images used in Training\end{tabular}} & \multicolumn{2}{c|}{\begin{tabular}[c]{@{}c@{}}Mean Absolute Loss (Absolute Loss\\ Standard Deviation, $r^2$)\end{tabular}}                                             \\ \cline{2-3} 
                                                                                                                                                                                                & \multicolumn{1}{c|}{\begin{tabular}[c]{@{}c@{}}Tested on 100 \\ real images\end{tabular}} & \begin{tabular}[c]{@{}c@{}}Tested on\\ remaining\\ real images\end{tabular} \\ \hline
3 genotypes (9, 32) + 32                                                                                                                                                                        & \multicolumn{1}{c|}{1.39 (1.66, 0.83)}                                                    & 1.62 (2.10, 0.81)                                                           \\ \hline
6 genotypes (17, 63) + 63                                                                                                                                                                       & \multicolumn{1}{c|}{1.41 (1.68, 0.82)}                                                    & 1.50 (1.95, 0.84)                                                           \\ \hline
9 genotypes (26, 80) + 80                                                                                                                                                                       & \multicolumn{1}{c|}{1.17 (1.32, 0.88)}                                                    & 1.34 (1.79, 0.87)                                                           \\ \hline
12 genotypes (34, 101) + 101                                                                                                                                                                    & \multicolumn{1}{c|}{1.11 (1.38, 0.88)}                                                    & 1.31 (1.78, 0.87)                                                           \\ \hline
15 genotypes (43, 123) + 123                                                                                                                                                                    & \multicolumn{1}{c|}{1.19 (1.59, 0.86)}                                                    & 1.31 (1.81, 0.87)                                                           \\ \hline
18 genotypes (50, 146) + 146                                                                                                                                                                    & \multicolumn{1}{c|}{0.95 (1.29, 0.90)}                                                    & 1.02 (1.56, 0.89)                                                           \\ \hline
21 genotypes (59, 165) + 165                                                                                                                                                                    & \multicolumn{1}{c|}{0.9 (1.33, 0.90)}                                                     & 0.99 (1.60, 0.90)                                                           \\ \hline
24 genotypes (66, 184) + 184                                                                                                                                                                    & \multicolumn{1}{c|}{0.86 (1.16, 0.92)}                                                    & 0.93 (1.66, 0.90)                                                           \\ \hline
27 genotypes (73, 200) + 200                                                                                                                                                                    & \multicolumn{1}{c|}{0.91 (1.28, 0.91)}                                                    & 1.02 (1.76, 0.89)                                                           \\ \hline
30 genotypes (80, 215) + 215                                                                                                                                                                    & \multicolumn{1}{c|}{0.87 (1.22, 0.92)}                                                    & 1.04 (1.74, 0.89)                                                           \\ \hline
33 genotypes (89, 236) + 236                                                                                                                                                                    & \multicolumn{1}{c|}{0.91 (1.30, 0.91)}                                                    & 1.08 (1.87, 0.89)                                                           \\ \hline
36 genotypes (96, 254) + 254                                                                                                                                                                    & \multicolumn{1}{c|}{0.83 (1.24, 0.92)}                                                    & 1.12 (2.00, 0.89)                                                           \\ \hline
39 genotypes (104, 285) + 285                                                                                                                                                                   & \multicolumn{1}{c|}{0.89 (1.34, 0.90)}                                                    &                                                                             \\ \hline
\end{tabular}%
}
\caption{The mean absolute loss and the standard deviation for the inflorescence branch counting task in canola with training on equal amount of real images and synthetic images from $C^S_4$ combined, and testing on 100 real images and remaining real images.}
\label{tab_4}
\end{table*}
\begin{table*}[]
\centering
\resizebox{\linewidth}{!}{%
\begin{tabular}{|c|cccccc|}
\hline
\multirow{2}{*}{\begin{tabular}[c]{@{}c@{}}Numbers of real plant genotypes\\ (Number of real plants,\\ Number of real images)\\ used in Training\end{tabular}} & \multicolumn{6}{c|}{Mean Absolute Loss (Absolute Loss Standard Deviation, $r^2$), tested on all remaining real images}                                                                                                                                                                                                                                                                                                                                                                                                                                                                                                                                                                                         \\ \cline{2-7} 
                                                                                                                                                               & \multicolumn{1}{c|}{\begin{tabular}[c]{@{}c@{}}No synthetic data \\ augmented for training\end{tabular}} & \multicolumn{1}{c|}{\begin{tabular}[c]{@{}c@{}}Synthetic data\\$C^S_1$ augmented\\ for training\end{tabular}} & \multicolumn{1}{c|}{\begin{tabular}[c]{@{}c@{}}Synthetic data\\$C^S_2$ augmented\\ for training\end{tabular}} & \multicolumn{1}{c|}{\begin{tabular}[c]{@{}c@{}}Synthetic data\\$C^S_3$ augmented\\ for training\end{tabular}} & \multicolumn{1}{c|}{\begin{tabular}[c]{@{}c@{}}Synthetic data\\$C^S_4$ augmented\\ for training\end{tabular}} & \begin{tabular}[c]{@{}c@{}}Synthetic data\\$C^S_5$ augmented\\ for training\end{tabular} \\ \hline
0 real data                                                                                                                                                    & \multicolumn{1}{c|}{}                                                                                    & \multicolumn{1}{c|}{12.22 (8.73, -4.29)}                                                                             & \multicolumn{1}{c|}{16.0 (6.06, -5.86)}                                                                               & \multicolumn{1}{c|}{2.15 (2.62, 0.73)}                                                                                & \multicolumn{1}{c|}{1.74 (2.08, 0.82)}                                                                                & 5.72 (3.88, -0.12)                                                                               \\ \hline
3 genotypes (9, 32)                                                                                                                                            & \multicolumn{1}{c|}{1.48 (1.72, 0.86)}                                                                   & \multicolumn{1}{c|}{5.67 (6.62, -0.97)}                                                                              & \multicolumn{1}{c|}{3.08 (3.78, 0.38)}                                                                                & \multicolumn{1}{c|}{1.86 (2.31, 0.77)}                                                                                & \multicolumn{1}{c|}{2.30 (2.45, 0.70)}                                                                                & 2.53 (2.81, 0.62)                                                                                \\ \hline
6 genotypes (17, 63)                                                                                                                                           & \multicolumn{1}{c|}{1.4 (1.72, 0.87)}                                                                    & \multicolumn{1}{c|}{3.34 (4.13, 0.26)}                                                                               & \multicolumn{1}{c|}{3.86 (4.43, 0.09)}                                                                                & \multicolumn{1}{c|}{1.51 (2.05, 0.83)}                                                                                & \multicolumn{1}{c|}{1.58 (1.97, 0.83)}                                                                                & 2.02 (2.54, 0.72)                                                                                \\ \hline
9 genotypes (26, 80)                                                                                                                                           & \multicolumn{1}{c|}{1.30 (1.78, 0.87)}                                                                   & \multicolumn{1}{c|}{2.42 (3.37, 0.55)}                                                                               & \multicolumn{1}{c|}{1.88 (2.67, 0.72)}                                                                                & \multicolumn{1}{c|}{1.41 (1.75, 0.86)}                                                                                & \multicolumn{1}{c|}{1.40 (1.96, 0.85)}                                                                                & 2.03 (2.51, 0.73)                                                                                \\ \hline
12 genotypes (34, 101)                                                                                                                                         & \multicolumn{1}{c|}{1.15 (1.80, 0.88)}                                                                   & \multicolumn{1}{c|}{1.97 (2.96, 0.67)}                                                                               & \multicolumn{1}{c|}{1.62 (2.73, 0.73)}                                                                                & \multicolumn{1}{c|}{1.35 (1.83, 0.86)}                                                                                & \multicolumn{1}{c|}{1.45 (1.89, 0.85)}                                                                                & 1.42 (2.11, 0.83)                                                                                \\ \hline
15 genotypes (43, 123)                                                                                                                                         & \multicolumn{1}{c|}{1.16 (1.64, 0.89)}                                                                   & \multicolumn{1}{c|}{1.60 (2.76, 0.74)}                                                                               & \multicolumn{1}{c|}{1.54 (2.31, 0.80)}                                                                                & \multicolumn{1}{c|}{1.11 (1.68, 0.89)}                                                                                & \multicolumn{1}{c|}{1.0 (1.70, 0.90)}                                                                                 & 1.82 (2.56, 0.75)                                                                                \\ \hline
18 genotypes (50, 146)                                                                                                                                         & \multicolumn{1}{c|}{1.02 (1.64, 0.88)}                                                                   & \multicolumn{1}{c|}{1.33 (2.24, 0.79)}                                                                               & \multicolumn{1}{c|}{1.49 (2.48, 0.75)}                                                                                & \multicolumn{1}{c|}{1.10 (1.76, 0.87)}                                                                                & \multicolumn{1}{c|}{1.11 (1.68, 0.87)}                                                                                & 1.46 (2.19, 0.79)                                                                                \\ \hline
21 genotypes (59, 165)                                                                                                                                         & \multicolumn{1}{c|}{0.94 (1.66, 0.89)}                                                                   & \multicolumn{1}{c|}{1.89 (2.92, 0.66)}                                                                               & \multicolumn{1}{c|}{1.73 (2.52, 0.73)}                                                                                & \multicolumn{1}{c|}{1.05 (1.74, 0.88)}                                                                                & \multicolumn{1}{c|}{1.08 (1.72, 0.88)}                                                                                & 1.33 (2.12, 0.82)                                                                                \\ \hline
24 genotypes (66, 184)                                                                                                                                         & \multicolumn{1}{c|}{1 (1.72, 0.89)}                                                                      & \multicolumn{1}{c|}{1.70 (2.56, 0.74)}                                                                               & \multicolumn{1}{c|}{1.36 (2.20, 0.81)}                                                                                & \multicolumn{1}{c|}{1.01 (1.74, 0.89)}                                                                                & \multicolumn{1}{c|}{1.14 (1.85, 0.87)}                                                                                & 1.19 (2.12, 0.84)                                                                                \\ \hline
27 genotypes (73, 200)                                                                                                                                         & \multicolumn{1}{c|}{1.02 (1.94, 0.87)}                                                                   & \multicolumn{1}{c|}{1.33 (2.46, 0.79)}                                                                               & \multicolumn{1}{c|}{1.12 (2.18, 0.84)}                                                                                & \multicolumn{1}{c|}{1.0 (1.87, 0.88)}                                                                                 & \multicolumn{1}{c|}{0.96 (1.62, 0.90)}                                                                                & 1.18 (2.03, 0.85)                                                                                \\ \hline
30 genotypes (80, 215)                                                                                                                                         & \multicolumn{1}{c|}{0.93 (1.75, 0.90)}                                                                   & \multicolumn{1}{c|}{1.14 (2.31, 0.83)}                                                                               & \multicolumn{1}{c|}{1.2 (2.26, 0.83)}                                                                                 & \multicolumn{1}{c|}{1.1 (1.82, 0.88)}                                                                                 & \multicolumn{1}{c|}{1.17 (1.92, 0.87)}                                                                                & 1.22 (2.18, 0.84)                                                                                \\ \hline
33 genotypes (89, 236)                                                                                                                                         & \multicolumn{1}{c|}{1.01 (1.96, 0.89)}                                                                   & \multicolumn{1}{c|}{1.30 (2.38, 0.83)}                                                                               & \multicolumn{1}{c|}{1.07 (1.91, 0.89)}                                                                                & \multicolumn{1}{c|}{1.18 (2.00, 0.87)}                                                                                & \multicolumn{1}{c|}{0.95 (1.70, 0.91)}                                                                                & 1.12 (2.13, 0.86)                                                                                \\ \hline
36 genotypes (96, 254)                                                                                                                                         & \multicolumn{1}{c|}{1.08 (1.94, 0.89)}                                                                   & \multicolumn{1}{c|}{1.37 (2.55, 0.82)}                                                                               & \multicolumn{1}{c|}{1.09 (2.27, 0.86)}                                                                                & \multicolumn{1}{c|}{1.29 (2.06, 0.87)}                                                                                & \multicolumn{1}{c|}{1.02 (1.85, 0.90)}                                                                                & 1.21 (2.26, 0.86)                                                                                \\ \hline
\end{tabular}%
}
\caption{Each table cell contains the mean absolute loss (respectively standard deviation and $r^2$) for the inflorescence branch counting task in canola. Training was done on real images as indexed in the first column, column 2 is where it was not augmented by any synthetic data, and column 3 through 7 are augmented by synthetic data from  $C^S_1, C^S_2, C^S_3, C^S_4, C^S_5$ respectively.}
\label{tab_3}
\end{table*}

\begin{figure*}[!h]
\setcounter{figure}{0}
  \centering
  \subfloat[]{\includegraphics[width=.5\linewidth]{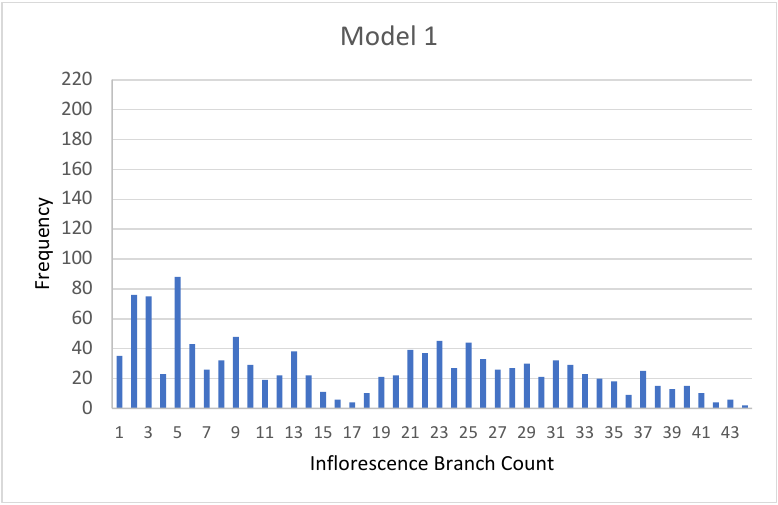}}
  \subfloat[]{\includegraphics[width=.5\linewidth]{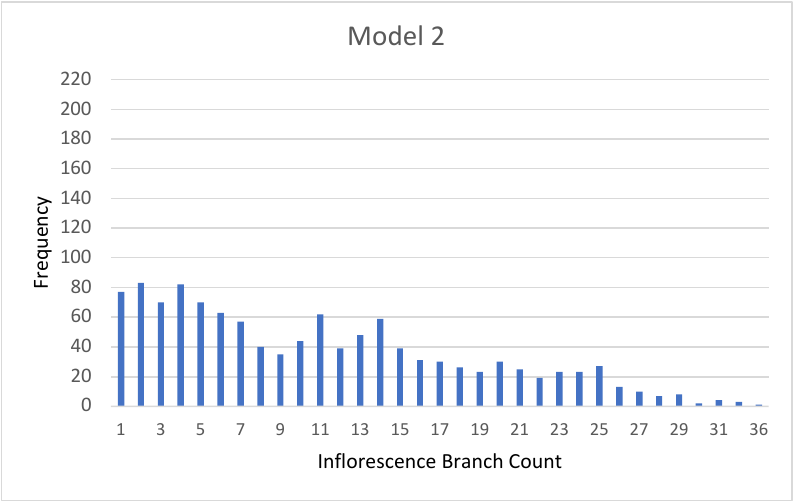}}  \hfill
  \subfloat[]{\includegraphics[width=.5\linewidth]{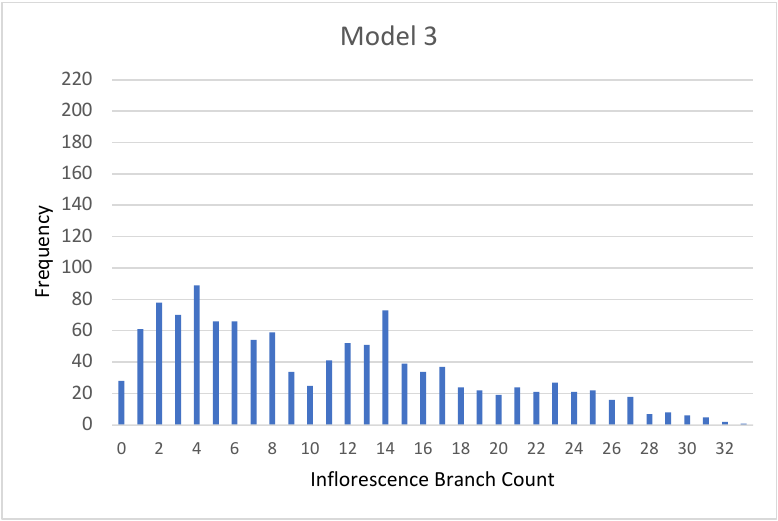}}
  \subfloat[]{\includegraphics[width=.5\linewidth]{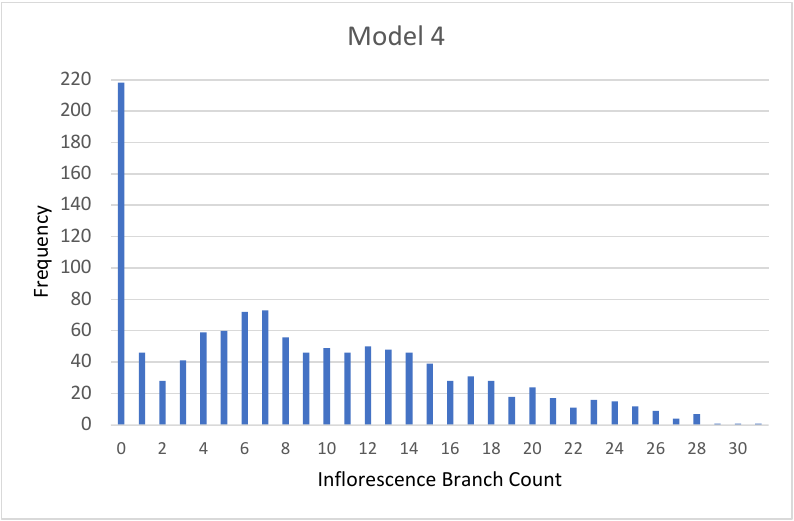}}
  \caption{Distribution of inflorescence branch count in synthetic canola images changes from first model using $C^S_1$ to the fourth model using $C^S_4$, which helped to improve results.}
  \label{fig_23}
\end{figure*}

\begin{figure*}[]
  \centering
  \subfloat[]{\includegraphics[width=.5\linewidth]{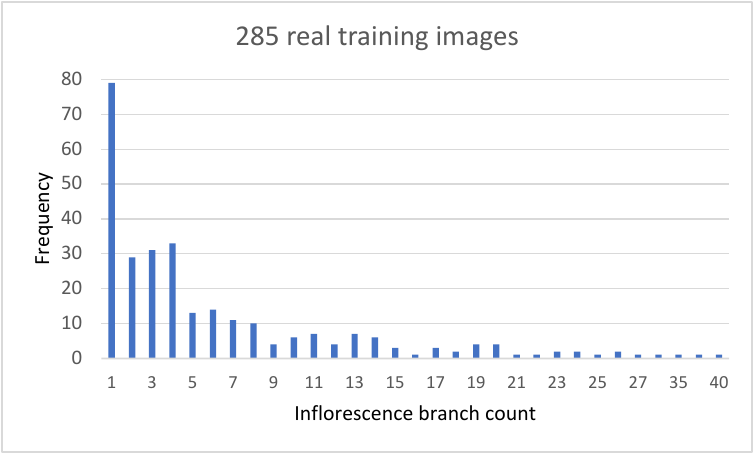}}
  \subfloat[]{\includegraphics[width=.5\linewidth]{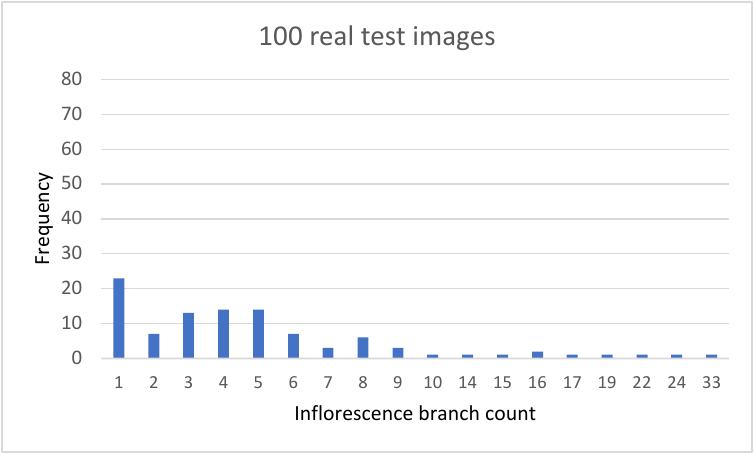}}
  \caption{Distribution of inflorescence branch count in real canola images used for training (left), and for testing (right) respectively.}
  \label{fig_24}
  \end{figure*}